%% file: main_conference.tex
\documentclass[twoside]{article}
\usepackage[preprint]{aistats2026}
\input{math_commands.tex}

\usepackage{subcaption}

\usepackage[round]{natbib}
\renewcommand{\bibname}{References}
\renewcommand{\bibsection}{\subsubsection*{\bibname}}

\begin{document}

\runningtitle{DPO with Unobserved Preference Heterogeneity: The Necessity of Ternary Preferences}

\twocolumn[

\aistatstitle{Direct Preference Optimization with Unobserved Preference Heterogeneity: The Necessity of Ternary Preferences}

\aistatsauthor{ Keertana Chidambaram \And Karthik Vinary Seetharaman \And  Vasilis Syrgkanis }

\aistatsaddress{ Stanford University } ]

\begin{abstract}
  \input{sections/abstract}
\end{abstract}

\input{sections/introduction}

\input{sections/background}

\input{sections/method}

\input{sections/identification}

\input{sections/experiments}

\bibliographystyle{plainnat}
\bibliography{main}

\input{sections/checklist}
\appendix
\onecolumn
\aistatstitle{APPENDIX}
\input{sections/appendix_algorithm}
\input{sections/appendix_identification}
\input{sections/appendix_experiment}

\end{document}

%% file: math_commands.tex
\usepackage[T1]{fontenc}
\usepackage{inputenc}
\usepackage{amssymb}
\usepackage{amsthm}
\usepackage{amsbsy}
\usepackage{mathtools}
\usepackage{esint}
\usepackage{bbm}
\usepackage{nicefrac}
\usepackage{graphicx}
\usepackage{float}
\usepackage{subcaption}
\usepackage{tcolorbox}
\usepackage{algorithm}
\usepackage{algpseudocode}
\usepackage{comment}
\usepackage{xcolor}
\usepackage{setspace}
\usepackage{color-edits}
\addauthor{kc}{magenta}
\addauthor{vs}{red}
\addauthor{hl}{red}
\addauthor{kf}{blue}
\usepackage{url}
\usepackage[breaklinks]{hyperref}
\usepackage{cleveref}
\usepackage{prettyref}
\usepackage{ifthen}
\usepackage[title]{appendix}
\definecolor{mySageGreen}{RGB}{188, 184, 138}
\theoremstyle{plain}
\newtheorem{theorem}{Theorem}[section]

\newtheorem{lemma}[theorem]{Lemma}

\theoremstyle{definition}

\newtheorem{assumption}{Assumption}

\theoremstyle{remark}

\makeatletter
\newtheorem*{rep@theorem}{\rep@title}
\newcommand{\newreptheorem}[2]{%
  \newenvironment{rep#1}[1]{%
    \def\rep@title{#2 \ref{##1}}%
    \begin{rep@theorem}}%
  {\end{rep@theorem}}}
\makeatother

\newreptheorem{theorem}{Theorem}
\newreptheorem{corollary}{Corollary}
\newreptheorem{assumption}{Assumption}

\crefname{theorem}{Theorem}{Theorems}
\crefname{lemma}{Lemma}{Lemmas}
\crefname{corollary}{Corollary}{Corollaries}
\crefname{proposition}{Proposition}{Propositions}
\crefname{conjecture}{Conjecture}{Conjectures}
\crefname{definition}{Definition}{Definitions}
\crefname{example}{Example}{Examples}
\crefname{remark}{Remark}{Remarks}
\crefname{assumption}{Assumption}{Assumptions}
\crefname{equation}{Equation}{Equations}
\crefname{algorithm}{Algorithm}{Algorithms}
\crefname{appendix}{Appendix}{Appendices}

\newcommand{\E}{\mathbb{E}}

\DeclareMathOperator*{\argmin}{arg\,min}
\DeclareMathOperator*{\argmax}{arg\,max}

\newcommand{\kl}{\mathbb{D}_{\text{KL}}}

\def\ddefloop#1{\ifx\ddefloop#1\else\ddef{#1}\expandafter\ddefloop\fi}
\def\ddef#1{\expandafter\def\csname bb#1\endcsname{\ensuremath{\mathbb{#1}}}}
\ddefloop ABCDEFGHIJKLMNOPQRSTUVWXYZ\ddefloop

\def\ddef#1{\expandafter\def\csname b#1\endcsname{\ensuremath{\mathbf{#1}}}}
\ddefloop ABCDEFGHIJKLMNOPQRSTUVWXYZ\ddefloop

\def\ddef#1{\expandafter\def\csname c#1\endcsname{\ensuremath{\mathcal{#1}}}}
\ddefloop ABCDEFGHIJKLMNOPQRSTUVWXYZ\ddefloop

\def\ddef#1{\expandafter\def\csname h#1\endcsname{\ensuremath{\widehat{#1}}}}
\ddefloop ABCDEFGHIJKLMNOPQRSTUVWXYZ\ddefloop

\def\ddef#1{\expandafter\def\csname hc#1\endcsname{\ensuremath{\widehat{\mathcal{#1}}}}}
\ddefloop ABCDEFGHIJKLMNOPQRSTUVWXYZ\ddefloop

\def\ddef#1{\expandafter\def\csname t#1\endcsname{\ensuremath{\widetilde{#1}}}}
\ddefloop ABCDEFGHIJKLMNOPQRSTUVWXYZ\ddefloop

\def\ddef#1{\expandafter\def\csname tc#1\endcsname{\ensuremath{\widetilde{\mathcal{#1}}}}}
\ddefloop ABCDEFGHIJKLMNOPQRSTUVWXYZ\ddefloop

\newcommand{\kibitz}[2]{\ifnum\Comments=1{\color{#1}{#2}}\fi}

\newcommand{\ba}{\begin{array}}
\newcommand{\ea}{\end{array}}
\newcommand{\bs}{\begin{align}\begin{split}\nonumber}
\newcommand{\bsnumber}{\begin{align}\begin{split}}
\newcommand{\es}{\end{split}\end{align}}

\def\balign#1\ealign{\begin{align}#1\end{align}}
\def\balignat#1\ealign{\begin{alignat}#1\end{alignat}}
\def\bitemize#1\eitemize{\begin{itemize}#1\end{itemize}}
\def\benumerate#1\eenumerate{\begin{enumerate}#1\end{enumerate}}

\newenvironment{talign}
 {\csname align\endcsname}
 {\endalign}
\def\balignt#1\ealignt{\begin{talign}#1\end{talign}}

%% file: sections/abstract.tex
Reinforcement Learning from Human Feedback (RLHF) has become central to aligning large language models with human values, typically by first learning a reward model from preference data which is then used to update the model with reinforcement learning. Recent alternatives such as Direct Preference Optimization (DPO) simplify this pipeline by directly optimizing on preferences. However, both approaches often assume uniform annotator preferences and rely on binary comparisons, overlooking two key limitations: the diversity of human evaluators and the limitations of pairwise feedback. In this work, we address both these issues. First, we connect preference learning in RLHF with the econometrics literature and show that binary comparisons are insufficient for identifying latent user preferences from finite user data and infinite users, while (even incomplete) rankings over three or more responses ensure identifiability. Second, we introduce methods to incorporate heterogeneous preferences into alignment algorithms. We develop an Expectation-Maximization adaptation of DPO that discovers latent annotator types and trains a mixture of LLMs accordingly. Then we propose an aggregation algorithm using a min-max regret fairness criterion to produce a single generative policy with equitable performance guarantees. Together, these contributions establish a theoretical and algorithmic framework for fairness and personalization for diverse users in generative model alignment.

%% file: sections/introduction.tex
\section{Introduction}
Reinforcement Learning from Human Feedback (RLHF) has emerged as a leading approach to align language models (LMs) with human preferences by learning a single reward model from preference data and using it to fine-tune the LM (\cite{ouyang2022training, stiennon2020learning, wang2023aligning}). Direct Preference Optimization (DPO) (\cite{rafailov2023direct}) sidesteps the reinforcement learning step by directly optimizing policies from preference comparisons, but like RLHF, it implicitly assumes a single reward model and, hence, homogeneous preferences in the target population. Therefore, these methods run the risk of aligning only to the preferences of majority groups and neglecting underrepresented groups with heterogeneous preferences, leading to suboptimal behavior for several parts of the population and to bias and discrimination. 

Existing attempts to address this issue typically learn more expressive reward models and then perform RL (e.g., with PPO), but DPO offers advantages in stability and simplicity by avoiding explicit reward modeling. For instance, (\cite{zhou2023beyond}) builds on DPO to implicitly learn a multi-objective reward model, and other work generalizes to multi-dimensional rewards (\cite{wang2024arithmetic, zhou2023beyond}), yet such approaches face two challenges: (i) they often require annotators to provide multi-dimensional ratings (e.g., safety, accuracy), which are more costly and harder to obtain than binary preference data (\cite{casper2023open}); and (ii) the objectives must be fixed before data collection, which is problematic since many latent cultural, political, or geographical factors shape preferences in ways that are difficult to anticipate (\cite{siththaranjan2023distributional, bai2022training}). 

The main contributions of this paper are summarized as follows:
\begin{itemize}
\item We introduce Expectation-Maximization Direct Preference Optimization (\textbf{EM-DPO}), a novel clustering algorithm using expectation–maximization that simultaneously uncovers latent user preference types and trains an ensemble of LLMs, where each element in the ensemble is tailored to each type directly using preference data. By discovering hidden patterns in user preferences and learning separate models for distinct preference groups, EM-DPO enables genuine personalization of LLMs to diverse user populations.
\item We propose MinMax Regret Aggregation (\textbf{MMRA}), an aggregation algorithm that combines the LLM ensembles learned with EM-DPO based on a min–max regret fairness criterion. This enables robust deployment for pluralistic alignment when individual user types are unknown at inference time, ensuring no preference group is severely underserved.
\item We establish a fundamental connection between preference learning in LLMs and the econometrics literature, revealing when identification of heterogeneous preferences is provably achievable under a simple linear reward model. We uncover that latent heterogeneous preferences are \textit{not identifiable} when each user provides a binary comparison, but they \textit{become identifiable}  when each user reports a single data point of their preferred response among three options. This identification result has profound implications for LLM personalization, as it demonstrates that the standard binary comparison paradigm is fundamentally insufficient for learning diverse user preferences, regardless of dataset size. Albeit the solution is simple and surprising; ask the user to choose among three options, instead of just two. We also validate this theory with empirical evidence proving that ternary preferences exhibit stronger performance than binary preferences.
\end{itemize}

The research contributions closest to our work are \cite{chakraborty2024maxmin} and \cite{park2024rlhf}. Our approach offers several key advantages: First, EM-DPO directly performs expectation-maximization, which provides better theoretical guarantees than the hard-clustering algorithms proposed by \cite{chakraborty2024maxmin} and \cite{park2024rlhf}. Second, motivated by our identification results demonstrating that multi-item preferences are necessary for learning heterogeneous user types, EM-DPO is specifically designed to handle multi-item comparisons in addition to binary preferences, a capability absent in prior work. Third, both our algorithms are reward-free similar to \cite{park2024rlhf}, avoiding the need for explicit reward modeling. Finally, our fairness criterion based on min-max regret differs fundamentally from \cite{chakraborty2024maxmin}, which uses max-min rewards. We next discuss further related work on RLHF with preference heterogeneity and defer further comparisons to related work on reward modeling and preference-based reinforcement learning to the Appendix.

\begin{figure*}[t]  
    \centering
    \includegraphics[width=0.9\textwidth]{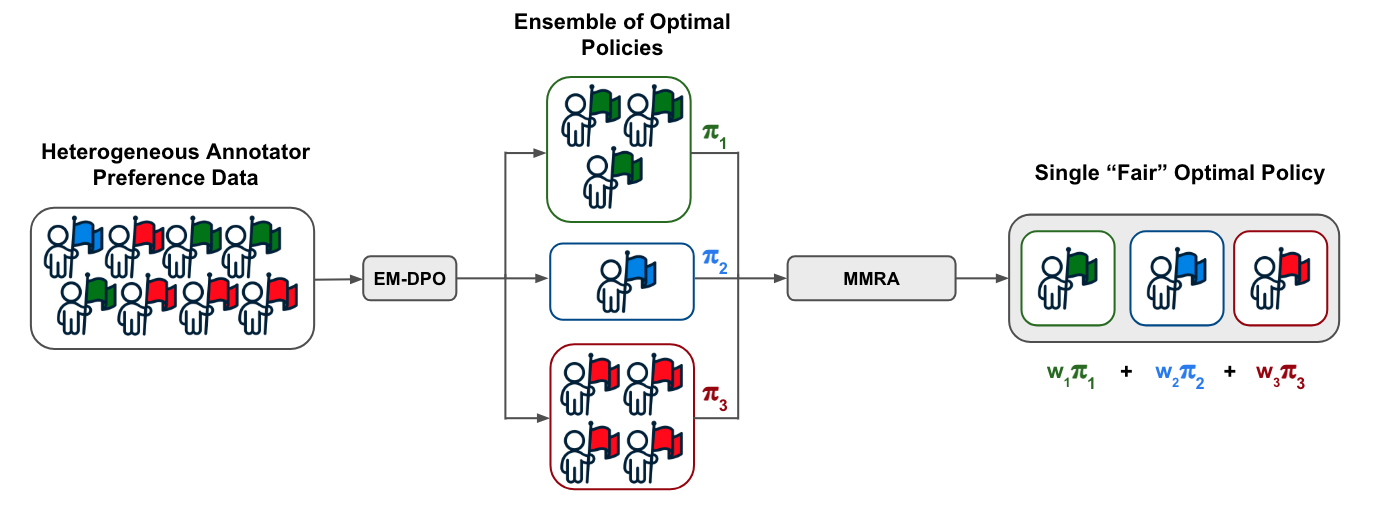}
    \caption{Proposed pipeline for learning an equitable policy. Step 1: Collect binary preferences from heterogeneous annotators. Step 2: Use EM-DPO to cluster annotators and derive an ensemble of optimal policies. Step 3: Apply MMRA to combine these policies into a single fair policy.}
    \label{fig:pipeline}
\end{figure*}

\textbf{RLHF With Diverse Preferences.} Diversity in annotator preferences has been recognized as a chief issue in RLHF \cite{dumoulin2023density}. Several studies have tried to solve the diverse population problem by learning more expressive reward functions and then using them to perform RLHF. For example, \cite{rame2024rewarded, jang2023personalized, chakraborty2024maxmin} maintains and learns several reward models at once. Similarly, \cite{wang2024arithmetic} learns a multi-dimensional reward model where each dimension provides rewards based on a different objective such as safety or usefulness. \cite{yang2024metaaligner} proposes a policy-agnostic method to perform multi-objective LLM alighment. Alternatively, \cite{siththaranjan2023distributional, li2024aligning} learns a distribution over fixed reward models. Finally, these reward models are combined using various strategies \cite{bakker2022fine, jang2023personalized, rame2024rewarded} to get a final reward model which is then used to perform RLHF. \cite{chakraborty2024maxmin} also learns multiple reward models, but performs RL by maximizing the minimum reward thereby ensuring that the final model is fair. The paper draws on elements of social choice theory, which \cite{conitzer2024social} argues is an effective path forward for RLHF research in general, specifically regarding issues with aggregating preferences. \cite{dai2024mapping} outlines a correspondence between the key principles and desiderata of social choice into the RLHF context.
In an orthogonal approach, \cite{zhong2024provable} utilizes meta-learning to learn diverse preferences. In general, trying to do RLHF with many reward models becomes expensive, making extending DPO \cite{rafailov2023direct} an attractive alternative. \cite{swamy2024minimaximalist} proposes SPO to sidestep reinforcement learning using the concept of a minimax winner from social choice theory, but only in the case of homogeneous preferences. \cite{rame2024warm} also deals with the idea of aggregating reward models to increase robustness. 
We instead propose a complete pipeline to learn one equitable policy for a heterogeneous population without appealing to reward model estimation at all.

%% file: sections/background.tex
\section{Background}\label{background}

The RLHF (\cite{ziegler2019fine, stiennon2020learning, ouyang2022training}) pipeline has two main inputs. The first is a language model, denoted as $\pi_{\text{SFT}}$, which is pre-trained on large-scale data and then fine-tuned using supervised learning. The second input is a static annotator preference dataset, $\mathcal{D} = \{x, y_w, y_l, h\}$, collected as follows: for a given prompt $x$, pairs of responses $(y_1, y_2)$ are generated from $\pi_{\text{SFT}}(\cdot|x)$, and a human annotator $h \in \mathcal{H}$ selects the preferred response. The winning and losing responses are denoted $y_w$ and $y_l$, respectively.

To model the ground truth of annotator choices, a common assumption links the observed preferences to a latent reward function via the Bradley–Terry–Luce model (\cite{bradley1952rank, ouyang2022training, rafailov2023direct}). Let $r^*(x,y)$ denote the true reward function. Then, for any pair $(y_1, y_2)$:
\begin{align}\label{bt_model}
\begin{split}
p_{r^*}(y_1 \succ y_2 | x) 
&= \frac{\exp(r^*(x, y_1))}{\exp(r^*(x, y_1)) + \exp(r^*(x, y_2))}\\
&= \sigma(r^*(x, y_1) - r^*(x, y_2))
\end{split}
\end{align}
The first step of RLHF fits a reward model $r_{\psi}(x,y)$ to approximate $r^*$ by minimizing the log-likelihood over the observed preferences:
\begin{align*}
\mathcal{L}(r_{\psi}; \mathcal{D}) = -\E_{\mathcal{D}}[\sigma(r_{\psi}(x,y_1) - r_{\psi}(x,y_2))]
\end{align*}

The second step fine-tunes the language model using reinforcement learning (PPO \cite{schulman2017proximal}) to maximize expected reward while regularizing deviation from the supervised model:
\begin{align}\label{rlhf_pol_llhood}
\begin{split}
\pi^*_{\phi} =& \argmax_{\pi_{\phi}} \E_{x \sim \mathcal{D},\, y \sim \pi_{\phi}(y|x)}
\big[ r_{\psi}(y, x) \big] \\
&- \beta \kl[\pi_{\phi}(y|x) \,\|\, \pi_{\text{SFT}}(y|x)]
\end{split}
\end{align}

Direct Preference Optimization (DPO) (\cite{rafailov2023direct}) bypasses the intermediate reward model by directly optimizing the policy using the preference dataset. Combining the Bradley–Terry model with the KL-regularized objective, DPO minimizes:
\begin{align*}
&\mathcal{L}(\pi_{\phi}; \pi_{\text{SFT}}, \mathcal{D}, \beta) \\
&= -\E_{\mathcal{D}} \left[ \log \sigma \left( \beta \log \frac{\pi_{\phi}(y_w|x)}{\pi_{\text{SFT}}(y_w|x)} - \beta \log \frac{\pi_{\phi}(y_l|x)}{\pi_{\text{SFT}}(y_l|x)} \right) \right] \\
&\pi_{\phi}^* = \argmin_{\pi_{\phi}} \mathcal{L}(\pi_{\phi}; \pi_{\text{SFT}}, \mathcal{D}, \beta)
\end{align*}

This framework forms the foundation for our extension to diverse annotators, which we address in the following sections.

%% file: sections/method.tex
\section{DPO for Diverse Annotators}

Both Reinforcement Learning from Human Feedback (RLHF) and Direct Preference Optimization (DPO) assume uniform preferences across the population and learn a single reward model ($r_{\psi}(y,x)$) either implicitly or explicitly. However, human preferences and values are inherently diverse. Consequently, RLHF and DPO tend to align with the majority opinion among annotators, introducing bias and potentially marginalizing minority perspectives. To mitigate this issue, we propose a pipeline consisting of two algorithms: Expectation-Maximization DPO (EM-DPO) for soft-eclustering diverse preference distributions and learning the optimally aligned policy for each cluster and Min-Max  DPO, which fairly aggregates the learned policies to minimize worst-case regret for any sub-group of annotators.

We do not impose structural assumptions on how preference data varies across annotators, instead modeling this variation as arising from an unobserved latent factor. We assume only that the latent factors are discrete and finite, without access to any observable indicators such as group labels for annotators. To handle this, we extend the expectation-maximization (EM) algorithm (\cite{dempster1977maximum, moon1996expectation}) to our setting, leveraging its ability to handle mixture data. The resulting algorithm, EM-DPO, soft-clusters annotators based on their observed preference data and learns an optimal policy for each cluster. We first look at the data generating process under this setting and then derive the EM algorithm.

\paragraph{Data Generating Process}
Let $Z$ present an annotator's latent factor capturing unobserved heterogeneity. We make the following assumption that there are a finite number of latent variables similar to (\cite{chakraborty2024maxmin, ramesh2024group}):

\begin{assumption}[Finite Latent Types]\label{ass:finite1}
For all $Z$, $Z \in \{z_1,\ldots,z_K\}$, where $K \in \mathbb{N}$ is some finite value.
\end{assumption}
Note that the value of $k$ need not be fixed a priori and can be treated as a hyperparameter (see Appendix for hyper-parameter tuning). The true reward function of an annotator is thus $r^*(y, x; Z)$, where $x$ and $y$ denote the prompt and the response respectively. Suppose there are $n$ annotators, i.e. $|\mathcal{H}| = n$, and for simplicity assume $m$ preference data points per annotator. For each  annotator $h$ indexed by $i \in [n]$, the preference data is generated by first sampling a latent factor $Z_i$ followed by  $m$ observed preferences with $V_{i,j} = (x^{i,j}, y_w^{i,j}, Y_r^{i,j})$ for $1 \leq j \leq m$ conditioned on $Z_i$, where $y_w^{i,j}$ is the preferred response and $Y_r^{i,j}$ is the set of rejected response(s). 

In the context of LLMs, prompts $X_{ij}$ are randomly assigned to annotators,  ensuring no correlation between an annotator's preference type and the assigned prompt. Thus, prompts are equally likely across preference types. We formalize this observation in the following assumption:
\begin{assumption}[Un-correlated Contexts and Latent Preference Types]\label{ass:uncorr1}
For all $k,\ell\in [K]$:
\begin{align*}
    p(X_{ij}\mid Z_i=z_k; \theta) =~& p(X_{ij}\mid Z_i=z_\ell;\theta) \\:=~& \rho(X_{ij}) 
\end{align*}
\end{assumption}

\subsection{EM Algorithm}

We start off with an offline dataset of annotator-level preferences $\mathcal{D}$ and $k$ policies (LLMs). Let $\theta$ be the set of parameters that parametrize both the distribution of $Z$ and the distribution of $V$ conditioned on $Z$, i.e. $p(Z;\theta)$ and $p(V\mid Z; \theta)$. At time step $t$ of the algorithm, let $\theta_t$ be the candidate parameters of the algorithm. Now the EM algorithm can be succinctly written as:
\begin{align*}
\theta_{t+1} 
&= \argmax_{\theta} Q(\theta\mid \theta_t) \\
\text{where,  } Q(\theta\mid \theta_t) 
&= \E_{Z\sim p(\cdot\mid V,\theta_t)}\left[\log(p(V, Z\mid \theta))\right]
\end{align*}

Breaking this down, there are two-steps, the E-step which is computing $Q(\theta\mid \theta_t)$ and the M-step which is finding the optimal value $\argmax_{\theta} Q(\theta\mid \theta_t)$. We defer to the Appendix for the derivation of both the E and M steps, that give rise to Algorithm~\ref{em-algo}.

\begin{algorithm*}[t]
\caption{EM-DPO}\label{em-algo}
\begin{algorithmic}[1]
    \State \textbf{Input:} Preference dataset $\mathcal{D}$ from annotators $\mathcal{H}$, with $m_i$ demonstrations from annotator $i$
    \State \textbf{Input:} Supervised fine-tuned model $\pi_{\text{SFT}}$
    \State \textbf{Input:} $K$ copies of $\pi_{\text{SFT}}$, denoted $\pi_{\phi_0, z_k}$ for $k=1,\ldots,K$
    \State Initialize mixture weights $(\eta_{1,0}, \ldots, \eta_{K,0}) = (1/K, \ldots, 1/K)$
    \For{$t = 0, 1, \ldots, T$}
        \State \textbf{E-step:} For each annotator $i \in \mathcal{H}$, compute
        $$
            \gamma_{i,k} = \frac{\eta_{k,t} \prod_{j=1}^{m_i} P_{\phi_t}(y_w^{i,j} \succ Y_r^{i,j} \mid x^{i,j}, z_k)}
                                {\sum_{\ell=1}^K \eta_{\ell,t} \prod_{j=1}^{m_i} P_{\phi_t}(y_w^{i,j} \succ Y_r^{i,j} \mid x^{i,j}, z_\ell)}
        $$
        where $Y_r^{i,j}$ is the set of rejected items in demonstration $j$ for annotator $i$, and
        $$
            P_{\phi_t}(y_w \succ Y_r \mid x, z_k) = \frac{\exp\!\Big(\log \tfrac{\pi_{\phi_t, z_k}(y_w \mid x)}{\pi_{\text{SFT}}(y_w \mid x)}\Big)}
                                                {\sum_{y \in \{y_w\} \cup Y_r} \exp\!\Big(\log \tfrac{\pi_{\phi_t, z_k}(y \mid x)}{\pi_{\text{SFT}}(y \mid x)}\Big)}
        $$
        \State \textbf{M-step:} Update
        $$
            \eta_{k,t+1} = \frac{1}{|\mathcal{H}|} \sum_{i \in \mathcal{H}} \gamma_{i,k}, 
            \qquad
            \phi_{t+1} = \argmax_{\phi} \sum_{i \in \mathcal{H}} \sum_{k=1}^K \gamma_{i,k} \sum_{j=1}^{m_i} \log P_{\phi}(y_w^{i,j} \succ Y_r^{i,j} \mid x^{i,j}, z_k)
        $$
    \EndFor
    \State \textbf{Return:} Policies $\{\pi^*_{k} = \pi_{\phi_T, z_k}\}_{k=1}^K$ and annotator weights $\{\gamma_{i,k}\}_{i \in \mathcal{H}}$
\end{algorithmic}
\end{algorithm*}

Note that if we do not share parameters across the policies for each preference type $z$, i.e. we have separate parameters $\phi_z$ for each $z\in \{z_1,\ldots, z_K\}$, then the optimization in the final step of EM-DPO also decomposes into separate policy optimization problems for each preference type:
\begin{align*}
    \phi_{z_k} = \argmax_{\phi_{z_k}} 
        \sum_{i \in \mathcal{I}} \sum_{j=1}^{m_i} 
        \gamma_{i,k} \,
        \log\!\big(P_{\phi}(y_w^{i,j} \succ Y_r^{i,j} \mid x^{i,j}, z_k)\big)
\end{align*}
Note that the latter is simply a weighted version of the multi-item DPO formulation (\cite{chen2024softmax}), where each demonstration $V_{i,j}$, which corresponds to the $j$-th demonstrations from annotator $i$, is assigned weight $\gamma_{i,k}$ when optimizing the policy parameters for preference type $z_k$. Alternatively, some parameters can be shared across policies for each preference type, in which case the final optimization problem should be solved simultaneously via stochastic gradient descent over the joint parameters $\phi$.

\subsection{Fair Aggregation via Min-Max Regret}
So far, we have outlined how to train an ensemble of LLMs, each optimized for one of the $K$ preference sub-groups. We now turn to the problem of aggregating this ensemble into a single fair policy that balances group preferences by minimizing worst-case regret across sub-populations.

We assume that EM-DPO is able to correctly identify data belonging to each sub-group and outputs policies $\pi_{k}^*$s that optimize the true reward for each group, $r^*(y, x; z_k)$. For ease of notation let $\E_\pi[\cdot] := \E_{x \sim \mathcal{D},\, y \sim \pi(\cdot \mid x)}[\cdot]$, then:
\begin{align*}
 \pi_{k}^* = \argmax_{\pi} \E_{\pi}\left[r^*(y, x; z_k)\right] - \beta \mathbb{D}_{\text{KL}}\left(\pi(y|x) || \pi_{\text{SFT}}(y|x)\right)
\end{align*}

Define $R_k(\pi)$ as the expected regret for population $Z = z_k$, measured relative to its population-optimal policy $\pi_k^*$:
\begin{align}\label{regret_def}
    R_k(\pi) := \E_{\pi^*_k}\left[r^*(y, x; z_k)\right] - \E_{\pi}\left[r^*(y, x; z_k)\right]
\end{align}
This captures the loss in expected reward for subgroup $k$ when following $\pi$ instead of its optimal policy $\pi^*_k$. Our fairness criterion is to minimize the worst-case subgroup regret. Accordingly, we seek a policy $\pi \in \Pi$ that solves:
\begin{align} \label{minmax-reg-obj}
    &\pi_* = \argmin_{\pi \in \Pi} \max_{w\in \Delta^{K-1}}  \sum_k w_k ([R_k(\pi)]^+  + \beta \mathbb{D}_{KL}(\pi || \pi_{\text{SFT}})) \nonumber\\
    &\beta \mathbb{D}_{KL}(\pi || \pi_{\text{SFT}})) = E_{\pi}\left[\beta \log\frac{\pi(y|x)}{\pi_{\text{SFT}}(y|x)} \right]\\
    &R_k(\pi) = \E_{\pi^*_k}\left[\beta \log\frac{\pi^*_k(y|x)}{\pi_{\text{SFT}}(y|x)}\right] - \E_{\pi}\left[\beta \log\frac{\pi^*_k(y|x)}{\pi_{\text{SFT}}(y|x)}\right]\nonumber
\end{align}
where $[x]^+ = \max\{x, 0\}$, $\beta$ is the regularization parameter and $\phi$ parametrizes $\pi$. Ideally, this objective could be directly optimized, with the maximizing player using multiplicative weights updates and the minimizing player applying gradient descent. However, such an approach is computationally and memory intensive, as it requires continuous generation from $\pi$ and simultaneous access to all $K$ policies in order to compute the log probabilities needed for estimating the $R_k(\pi)$ terms. In this paper, we deploy a light-weight approximation that retains the multiplicative-weights–versus–gradient-descent structure, as outlined in Algorithm \ref{minmax-algo}. In Appendix, we also show that if one limits to affine combinations of the ensemble models, then solving the minimax problem can be performed efficiently with provable guarantees.

\begin{algorithm*}[t]
\caption{MMRA-LW (Lightweight)}\label{minmax-algo}
\begin{algorithmic}[1]
    \State \textbf{Input:} EM-DPO ensemble policies $\{\pi^*_1,..,\pi^*_K\}$ \& weights $\{\gamma_{i,k}: i \in \mathcal{H}, k \in [K]\}$; SFT policy ${\pi_{\text{SFT}}}$; preference dataset $\mathcal{D}$; learning rate $\eta$; regularization parameter $\beta$;
    \State \textbf{Initialize:} weights $(w_1^0,...,w_K^0) = (1/K,...,1/K)$; current policy $\pi^0 = \pi_{\text{SFT}}$ for $t=0$
    \For{$t = 0,1,\ldots,T$}
        \State Compute weights as $\gamma_i = \sum_k w_k^{t-1} \gamma_{i,k}$
        \For{a few batches in $\mathcal{D}$}
            \State Perform weighted DPO as in Algorithm \ref{em-algo} to get $\pi^t$
        \EndFor
        \State Compute regrets $R_k(\pi^t)$ as per \ref{minmax-reg-obj}
        \State Update weights $w_k^t \propto \exp(R_k(\pi) \times \eta)$
    \EndFor
\end{algorithmic}
\end{algorithm*}

%% file: sections/identification.tex
\section{Identifiability: The Value of Three Choices}\label{identification-proof}

A natural question arises: under what conditions can we reliably recover the preference distribution in the population from observed data? In particular, since we assume no access to annotator identities or group labels, we must determine whether the latent distribution over preference types is uniquely identifiable from preference observations alone. In this section, we formalize this identifiability question and perform the analysis under a linear reward model. Somewhat surprisingly, we find that binary preferences suffer from fundamental non-identifiability, while multi-item preferences, even incomplete ternary preferences, enable unique recovery of the latent preference distribution.

Building on the Bradley–Terry framework introduced in Section~\ref{background}, we consider a simplified reward model for the identification analysis in our paper where each annotator's reward function is linear in a known feature representation:
\begin{align*}
r^*(x, y \mid \beta) = \beta^\top \psi(x, y),
\end{align*}
where $\psi(x,y) \in \mathbb{R}^d$ encodes features of prompt-response pairs, and $\beta \in \mathbb{R}^d$ captures user-specific preferences and can vary across annotators. While the main motivation for this simplification was to enable analysis, this linear model also has practical relevance. For instance, imagine a scenario where responses are evaluated along interpretable axes such as relevance, informativeness, and style. Each annotator may weigh these axes differently; one user might prioritize relevance heavily, while another emphasizes style. Representing overall preference as a weighted sum of these features naturally leads to a linear reward function. This linear reward modeling approach is also widely used in the alignment literature, particularly in works that study multi-objective rewards (\cite{wang2024arithmetic, zhou2023beyond, yang2024rewards}). Also, note that while we focus on linear rewards for the identifiability analysis, the algorithms presented in this paper can be applied to arbitrary reward models. 

When the preference vector $\beta$ is drawn from a distribution $f(\beta)$, the aggregate choice probabilities correspond to the random coefficient logit model (\cite{boyd1980effect, cardell1980measuring}):
\begin{multline} \label{eq:rc-logit}
p_{r^*}(y_1 \succ y_2, ..., y_n | x) = \\\int \frac{\exp(\beta^\top \psi(x, y_1))}{\sum_{i=1}^{n} \exp(\beta^\top \psi(x, y_i))} f(\beta) d\beta
\end{multline}
In this setting, the primary goal is to uncover $f$, which is the distribution over $\beta$ across the population, since this distribution determines both aggregate choice probabilities and the range of individual user preferences. Identification means that the true distribution $f_0$ is the unique $f$ that solves this equation for all possible values of the feature vectors $\psi(x, y_i)$. Equivalently, for any two distinct distributions $f_0 \neq f_1$, there exists some configuration of features $(x, y)$ where the choice probabilities differ: $p_{r^*}(y_1 \succ y_2, ..., y_n | x; f_0) \neq p_{r^*}(y_1 \succ y_2, ..., y_n | x; f_1)$. Understanding the conditions under which this distribution can be uniquely identified is the focus of our subsequent analysis.

\paragraph{Case 1: Single Binary Preference per User, Infinite Users.}  This setting closely resembles real-world scenarios. In this case, the distribution over user parameters, $f(\beta)$, is not fully identifiable:

\begin{lemma} \label{binary_identification}
Under the random coefficient logit model~\eqref{eq:rc-logit} with binary preferences, $f$ is not identifiable.
\end{lemma}
\begin{proof}
Consider a distribution $f$ where, with probability $1/2$, the user parameter is $\beta$, and with probability $1/2$ it is $-\beta$. For any pair $(x, y_1, y_2)$, the aggregate preference is
\begin{align*}
p_{r^*}(y_1 \succ y_2 \mid x, f) &= 0.5\,\sigma(\beta^\top (\psi(x, y_1)-\psi(x, y_2))) \\
&+ 0.5\,\sigma(-\beta^\top (\psi(x, y_1)-\psi(x, y_2)))\\
&= 0.5
\end{align*}
where we used $\sigma(x)=1-\sigma(-x)$. Hence, multiple distinct distributions (here, $\beta$ vs. $-\beta$) yield the same aggregate probabilities, proving $f$ is not identifiable.
\end{proof}

\paragraph{Case 2: Single Ternary Preference per User, Infinite Users.}  
Comparisons over at least three items resolve the non-identifiability problem that arises in binary preferences, where even infinitely many users with finite observations per user cannot provably recover the underlying preference distribution. Under mild conditions, (\cite{fox2012random}) shows that the distribution $f(\beta)$ becomes uniquely identifiable with infinitely many users, at least one preference per user, and sufficient variability in the observed ternary (even incomplete) rankings. We re-state this theorem here for completeness, the complete proof can be found in \cite{fox2012random}:

\begin{theorem}[Identification of Random Coefficients Logit] \label{identification}
If the following conditions hold:
\begin{itemize}
    \item The absolute moments of $f$, given by $m_l = \int \|\beta\|^l f(\beta)\, d\beta$, are finite for $l \geq 1$ and satisfy the Carleman condition: $\sum_{l \geq 1} m_l^{-1/l} = \infty$.
    \item The feature vectors $\psi(x, y_i)$ for each alternative $i \in \{1, ..., n\}$ take on support in an open set containing $\psi(x, y_i) = 0$ for all $i$.
    \item $\beta$ is independent of the features $\psi(x, y_i)$.
    \item $n \geq 3$ (at least 3 alternatives in the choice set).
\end{itemize}
Then the density $f(\beta)$ is nonparametrically identified in the random coefficients logit model from equation~\eqref{eq:rc-logit}.
\end{theorem}

The assumptions presented in Theorem \ref{identification} have natural interpretations in the RLHF setting. The Carleman condition, satisfied by most common distributions (normal, uniform, exponential), ensures $f$ is uniquely determined by its moments by preventing them from growing too quickly. The assumption that feature vectors span an open set containing 0 requires the prompt-response dataset to exhibit sufficient variation along multiple dimensions (e.g., helpfulness, harmfulness, coherence) to disentangle different annotator preference types. Finally, the independence of $\beta$ and the feature vectors $\psi(x, y_i)$ extends Assumption \ref{ass:uncorr1}, which assumes prompts are independent of user type, to require that both prompts and responses shows to a user are independent of the user's type.

\paragraph{Case 3: Many Diverse Binary Preferences for Each User}  We also note that if we have many binary preference observations from a single user with sufficiently diverse comparisons across items, their parameter vector $\beta$ can be identified:

\begin{lemma}
Define the matrix $U \in \mathbb{R}^{m \times d}$ whose rows are $\phi(x, y_1) - \phi(x, y_2)$ across $m$ preference pairs. If $U$ is full rank, then
\begin{align*}
\beta^\top (\psi(x, y_1)-\psi(x, y_2)) = \log\frac{p_{r^*}(y_1 \succ y_2 | x, \beta)}{1 - p_{r^*}(y_1 \succ y_2 | x, \beta)}
\end{align*}
can be solved uniquely for $\beta$. 
\end{lemma}

However, this solution requires a substantially rich set of responses per user. In particular, the number of diverse binary preferences per user needs to scale with the number of parameters in the reward model. For realistic reward models, we expect the feature map $\psi$ to be very high-dimensional and, therefore, this solution to the non-identifiability problem would require a very rich set of responses per user, essentially growing to infinity, as we consider richer and richer reward models. 

%% file: sections/experiments.tex
\section{Experiments}

\begin{table*}[t]
\caption{Mean Reward Margins and Accuracies for Different Algorithms and User Types for MPI Dataset}
\label{mpi-results-table}
\begin{center}
\begin{tabular}{lcccccc}
\hline
& \multicolumn{3}{c}{\textbf{Reward Margins}} & \multicolumn{3}{c}{\textbf{Accuracies}} \\
\cline{2-4} \cline{5-7}
\textbf{Method} & \textbf{P1} & \textbf{P2} & \textbf{P3} & \textbf{P1} & \textbf{P2} & \textbf{P3} \\
\hline
True Label DPO & 0.977 & 1.226 & 0.696 & 0.760 & 0.773 & 0.710 \\
EMDPO Ternary & 0.152 & \textbf{0.423} & \textbf{0.729} & \textbf{0.584} & \textbf{0.584} & 0.634 \\
EMDPO Binary & 0.116 & 0.134 & 0.585 & 0.515 & 0.550 & 0.651 \\
Cluster DPO Ternary & \textbf{0.246} & 0.257 & 0.688 & 0.560 & 0.562 & 0.591 \\
Cluster DPO Binary & 0.222 & 0.223 & 0.546 & 0.544 & 0.547 &  0.584 \\
Vanilla DPO & 0.031 & 0.020 & 0.228 & 0.532 & 0.512 & \textbf{0.666} \\
\hline
\end{tabular}
\end{center}
\end{table*}

In this section, we provide empirical evidence to answer the questions: (1) Does the EM-DPO algorithm learn high-quality clusters? (2) How fair are the final policies learned by Min-Max Regret DPO? (3) Do ternary preferences out-perform binary preferences in the adversarial case we discussed in section \ref{binary_identification}.

\subsection{Results}
\begin{table*}[t]
\caption{Max Mean Reward Margins and Accuracies for Different Clustering Algorithms and User Types for Global Opinions Dataset}
\label{combined-results-table}
\begin{center}
\begin{tabular}{lcccccccc}
\hline
& \multicolumn{4}{c}{\textbf{Reward Margins}} & \multicolumn{4}{c}{\textbf{Accuracies}} \\
\cline{2-5} \cline{6-9}
\textbf{Method} & \textbf{BR} & \textbf{IN} & \textbf{MX} & \textbf{PK} & \textbf{BR} & \textbf{IN} & \textbf{MX} & \textbf{PK} \\
\hline
True Label DPO & 0.733 & 1.302 & 1.233 & 1.183 & 0.703 & 0.740 & 0.738 & 0.779 \\
EMDPO & \textbf{0.853} & \textbf{1.474} & \textbf{1.558} & \textbf{1.617} & \textbf{0.710} & 0.715 & 0.747 & \textbf{0.767} \\
Cluster DPO & 0.750 & 1.228 & 1.307 & 1.163 & 0.702 & 0.718 & 0.747 & 0.765 \\
Vanilla DPO & 0.613 & 1.139 & 1.187 & 1.187 & 0.681 & \textbf{0.732} & \textbf{0.754} & 0.754 \\
\hline
\end{tabular}
\end{center}
\end{table*}



\begin{table*}[t]
\caption{Regret Values for Each Latent Sub-group Across Datasets}
\label{combined-regrets-table}
\begin{center}
\begin{tabular}{lcccccccccc}
\hline
& \multicolumn{5}{c}{\textbf{Global Opinions}} & \multicolumn{5}{c}{\textbf{MPI}} \\
\cline{2-6} \cline{7-11}
\textbf{Algorithm} & \textbf{1} & \textbf{2} & \textbf{3} & \textbf{4} & \textbf{Max} & \textbf{1} & \textbf{2} & \textbf{3} & \textbf{4} & \textbf{Max} \\
\hline
MMRA-LW & 0 & 0 & 1.73 & 0 & \textbf{1.73} & 0 & 3.44 & 0 & 1.74 & \textbf{3.44} \\
Uniform Sampling & 2.84 & 2.61 & 3.54 & 0 & 3.54 & 5.98 & 6.26 & 4.44 & 4.80 & 6.26 \\
Vanilla DPO & 0.78 & 0 & 3.18 & 0 & 3.18 & 0 & 4.14 & 0 & 1.46 & 4.14 \\
\hline
\end{tabular}
\end{center}
\end{table*}

\paragraph{Datasets:}  
We evaluate on two datasets that contain subgroups with differing preferences and imbalanced sizes, making them well-suited for studying pluralistic alignment: (1) GlobalOpinionQA (\cite{durmus2023towards}), and (2) MPI (\cite{jiang2022mpi}).


\textbf{GlobalOpinionQA.} This dataset contains country-level polling data on politics, religion, economics, and related topics. For each question, annotators respond according to their country-specific distribution, after which a second option is randomly rejected. For example, a Mexican annotator answering ``Do you support or oppose using the army to fight drug traffickers?'' responds with probabilities of 84\% support, 13\% oppose, and 3\% refuse/don’t know. We focus on four countries: Britain (15\%), Indonesia (20\%), Mexico (30\%), and Pakistan (35\%), comprising a total of 48000 preference pairs. Each question is augmented with 10 GPT-4 generated rephrasings, yielding 11 variants per question (original plus 10 rephrasings). 

For training, eight variants are used and three are reserved for validation and testing. To construct a user, we sample 32 unique questions and assign a random rephrasing from the appropriate split. Using this procedure, we generate 1,500 users for training and 400 for testing. Binary preferences are simulated from annotators and used to run the binary preference variant of EMDPO.  

\textbf{MPI.} The MPI dataset contains 990 phrases annotated with trait scores in $\{-1,0,+1\}$ for one of the five OCEAN personality traits \cite{mccrae1992introduction}. For example, ``act wild and crazy’’ scores $+1$ on Extraversion, while ``readily overcome setbacks’’ scores $-1$ on Conscientiousness. We define three synthetic personalities as vectors in $\mathbb{R}^5$:  $P1 = (3, 0, 2, 0, -2.5), P2 = (-3, 0, -2, 0, 2.5), P3 = (0, 2, 0, 2, 0)$
sampled with probabilities $0.3$, $0.3$, and $0.4$. The relation $P2 = -P1$ creates an adversarial pair that is non-identifiable under binary preferences, as discussed in Section~\ref{identification-proof}. 

Phrase rewards are computed as the inner product of the personality vector with phrase trait scores. To generate data, we (i) sample a personality, (ii) select one of 50 paraphrases of the instruction ``Choose the option that resonates most with your personality,'' (iii) draw phrases from the MPI dataset, and (iv) simulate preferences using the Bradley Terry model with either two or three items. Each user contributes exactly one preference pair, preventing identifiability. Both binary and ternary preferences are simulated, and we run both versions of EMDPO for comparison.  

\paragraph{Methods:} Our approach consists of two components: clustering and aggregation. In addition to comparing with \texttt{Vanilla DPO}, which trains a single DPO model on the full dataset without clustering, we evaluate the following benchmarks: For clustering, our primary method is \texttt{EM-DPO}. For the MPI dataset, we run both the binary and ternary preference versions of \texttt{EM-DPO}. As a baseline, we use \texttt{Cluster DPO}, which partitions users with K-means clustering on response embeddings and then trains DPO separately on each cluster to produce an ensemble of policies. We also include \texttt{True Label DPO} as a benchmark, which trains an ensemble of models directly on large data from the the original labels without any clustering algorithm. This is in a way the "best possible" situation where we know the latent variables and have large data. 

For aggregation, our main method is \texttt{Min-Max Regret Aggregation}, which combines the ensemble of policies by minimizing the maximum regret across annotator groups inferred in the clustering step. We compare against \texttt{Uniform Sampling}, which averages the ensemble policies uniformly.

\paragraph{Metrics:} To evaluate the quality of the learned clusters, we measure how well the ensemble covers diverse user preferences. For each user group, we identify the best LLM in the ensemble, defined as the policy that achieves the highest average reward margin on that group’s evaluation dataset. This captures the extent to which clustering produces specialized policies aligned with different user populations, and it is invariant to the indexing of clusters since only the best match matters. Formally, if $\{\pi^*_{i}\}_{i=1}^K$ denotes the ensemble of policies returned by clustering and $\mathcal{D}_{k'}$ is the evaluation dataset for group $k'$, the metric is:
\begin{multline*}
\text{Max-mean reward margin}_{k'} \\
= \max_{i \in [K]} \E_{(y_w, y_l, x) \sim \mathcal{D}_{k'}} 
   \Bigg[ \beta \log \frac{\pi^*_{i}(y_w | x)/\pi_{\text{SFT}}(y_w | x)}
                          {\pi^*_{i}(y_l | x)/\pi_{\text{SFT}}(y_l | x)} \Bigg]
\end{multline*}

For the aggregation step, since we adopt the min–max regret objective as our fairness criterion, we evaluate the aggregated policy by measuring the worst-case regret across user sub-populations. This metric quantifies the maximum shortfall any sub-population experiences, ensuring that the evaluation emphasizes fairness across all groups:
\begin{align*}
    \text{Max-regret}_{k \in [K]} =\max_{k\in[K]} R_{k}(\pi) 
\end{align*}
where $R_k(.)$ is as defined in Eq. \ref{regret_def} and $\pi$ is the aggregated policy.

\paragraph{Discussion} For the GlobalOpinions dataset, EM-DPO achieves strong performance on reward margins compared to baselines and delivers comparable accuracy. Notably, it outperforms models trained on large amounts of data with true labels in terms of reward margins, suggesting that EM-DPO can uncover latent structure even within annotated labels. The MMRA algorithm also performs well, yielding zero positive regret for Britain, Indonesia, and Pakistan, and only modest regret for Mexico, relative to uniform sampling from the ensemble policies and just training a single policy using Vanilla DPO.

On the MPI dataset, combining ternary preferences with EM-DPO leads to the best performance across both reward margins and accuracy. Interestingly, even Cluster-DPO with ternary preferences (which clustering prompts and chosen text before applying DPO) surpasses EM-DPO trained with binary preferences. This highlights the value of ternary preferences, particularly in the adversarial setting we constructed. For the aggregation step, we also notice that MMRA-LW out performs uniform sampling and vanilla DPO significantly, almost halving the regret values.

\section{Limitations \& Open Questions}
First, we assume that annotators belong to one of $K$ discrete latent groups, whereas in reality preferences most likely lie on a continuous spectrum. However, many areas (psychometrics, economics, recommender systems) use discrete "types" as standard approximations to continuous preference spaces (\cite{hagenaars2002applied, train2009discrete, sarwar2001item}). Moreover, this assumption improves interpretability and facilitates identification with limited data. Second, our method relies on the expectation–maximization (EM) algorithm, which is sensitive to initialization and may converge to saddle points. While we initialize using assignment from the K-means cluster algorithm, alternative strategies such as leveraging annotator demographics or employing an LLM-as-a-judge for initial estimates could improve robustness. Understanding the convergence and stability of EM in the context of large language models is an interesting avenue for future research. Additionally, extending the identification theory for more general classes of reward models beyond linear rewards is an open question.

%% file: sections/appendix_algorithm.tex
\section{Additional Related Work}
\paragraph{DPO Generalizations: } Since DPO's inception \cite{rafailov2023direct}, there has been a growing line of literature on its generalizations, some of which we highlight here. \cite{le2024multi} generalizes DPO to the case of multiple SFT models, while \cite{zhou2024beyond} generalizes to multiple objectives. \cite{zeng2024token, rafailov2024r} work on extending DPO to work at the token level. \cite{wang2023beyond} extends DPO to work with other types of divergence terms, while \cite{wu2024towards} relates DPO to DRO in order to robustify it. \cite{badrinath2024hybrid} augments DPO with a computable advantage function to create a hybrid between DPO and RLHF.

\paragraph{Preference-Based Reinforcement Learning: }Reinforcement learning from preferences has been an active research area for some time, providing a way to train on tasks for which explicitly defining rewards is hard \cite{wirth2017survey, lee2021b, abdelkareem2022advances}.  In particular, \cite{christiano2017deep, ibarz2018reward} show that using human preferences to guide reinforcement learning (RLHF) is particularly effective on a variety of tasks, such as training robots. More recently, RLHF has become a very popular technique to fine-tune language models to do a variety of tasks such as summarization \cite{ouyang2022training, ziegler2019fine, stiennon2020learning, wu2021recursively}. RLHF has also been used to align language models \cite{bai2022training, askell2021general}. \cite{casper2023open} details several open problems in the field of RLHF, including those related to the feedback itself, particularly the inverse relation between richness and efficiency. Some work has been done on this problem with regards to language-based feedback in particular \cite{fu2019language, zhou2021inverse} as well as in more general settings \cite{hwang2024sequential}, but specific applications to LLMs have not been fully explored.

\paragraph{Challenges with Reward Modeling: } In general, human preferences can be difficult to represent using reward models \cite{hong2022sensitivity}, and the validity of reward modeling itself is still somewhat debated \cite{bowling2023settling, bobu2023aligning, skalse2022the}. Some work has also been done to take personality into account when reward modeling \cite{lindner2022humans, lee2021b}, but this area remains open. In general, taking human irrationality into account when reward modeling (to optimize a more accurate reward function) leads to a trade-off between efficiency and accuracy \cite{shah2019feasibility, nguyen2017reinforcement}. Work has been done on inverse RL with particular models of suboptimality such as myopia \cite{evans2016learning}, noise \cite{zheng2014robust}, and risk-sensitivity \cite{majumdar2017risk}, but dealing with general irrationalities remains open. A recent work by \cite{golz2025distortion} shows that under heterogeneous user preferences, standard alignment methods (PPO-based RLHF and DPO) built on the Bradley-Terry model can produce policies whose average utility falls short of the optimal achievable average utility. \cite{golz2025distortion} recommends instead to use Nash Learning from Human Feedback \cite{munos2024nash} as an efficient alternative to the Bradley Terry model.

\section{Algorithms}
\subsection{EM-DPO Derivation}
\subsubsection{M-Step}
First we parametrize the the objective of the M-step, $Q(\theta\mid \theta_t)$. We can break down the optimization objective as follows:
\begin{align} 
Q(\theta\mid \theta_t)
&= \E_{Z\sim p(\cdot\mid V,\theta_t)}\left[ \log\left(p(V, Z; \theta)\right)\right] \\
&= \E_{Z\sim p(\cdot\mid V,\theta_t)}\left[ \log\left(\prod_{i=1}^n p(V_i, Z_i; \theta)\right)\right] \\
&= \E_{Z\sim p(\cdot\mid V,\theta_t)}\left[\sum_{i=1}^n \log \left(p(V_i, Z_i; \theta) \right)\right] \\
&= \E_{Z\sim p(\cdot\mid V,\theta_t)}\left[\sum_{i=1}^n \log(p(V_i\mid Z_i; \theta)) + \log(p(Z_i;\theta))\right]\\
&= \E_{Z\sim p(\cdot\mid V,\theta_t)}\left[\sum_{i=1}^n \log(p(V_i\mid Z_i; \theta))\right] + 
\E_{Z\sim p(\cdot\mid V,\theta_t)}\left[\log(p(Z_i;\theta))\right]
\end{align}

For our problem, $\theta$ comprises of three sets of parameters: $\phi$ which parametrizes the policies, $\eta$ which parametrizes the latent distribution of the user-types $p(Z;\theta)$, and $\rho$ which parametrizes the distribution of the prompts $X$. We will first see how this parametrizes the optimization problem.

First we look at $\log(p(Z_i;\theta))$. The latent factors $Z$ take values from a discrete set of $K$ values $\{z_1,\ldots,z_K\}$. In this case, we can assume a fully non-parametric likelihood $p(Z;\theta)$, where $\eta_k = p(z_k;\theta) \in \Delta(K)$, the $K$-dimensional simplex. Further, we note that:
\begin{align}
p(Z_i ; \theta) = \sum_{k=1}^K \eta_k \mathbf{1}\{Z_i = z_k\}
\end{align}

Further, assuming that $p(V_i\mid Z_i;\theta)$ does not depend on the vector $\eta$, so that $p(V_i\mid Z_i;\theta)=p(V_i\mid Z_i;\phi,\rho)$, the original criterion decomposes into two separate optimization problems:


\begin{equation}\label{eqn:M-step}
    \begin{aligned}
    \eta_{t+1} =~& \argmax_{\eta} \E_{Z\sim  p(\cdot\mid V,\theta_t)}\left[\sum_{i=1}^n \log\left(\sum_{k=1}^K \eta_k 1\{Z_i=z_k\}\right)\right]\\
    \phi_{t+1} =~& \argmax_{\phi,\rho} \E_{Z\sim p(\cdot\mid V,\theta_t)}\left[\sum_{i=1}^n \log(p(V_i\mid Z_i; \phi,\rho))\right]
\end{aligned}
\end{equation}

Next, we look at the parametrization of $p(V_i\mid Z_i; \phi,\rho)$. Note that, in our situation, both the latent factors and observed variables $(Z_i, V_i)$ are independent across the $n$ annotators and therefore, the likelihood and the prior factorizes across the annotators. Moreover, conditional on the latent factor $Z_i$, the $V_{i,j}$ are independently distributed across $j$ and for each $j$ the conditional likelihood takes a logistic form, as follows:
\begin{align}
p(V_i\mid Z_i; \phi,\rho) 
=~& \prod_{j=1}^m p(V_{i,j}\mid Z_i;\phi,\rho) \\
=~&  \prod_{j=1}^m P(y_w^{i,j} \succ Y_r^{i,j}, X^{i,j} \mid Z_i; \phi,\rho)\\
=~& \prod_{j=1}^m P(y_w^{i,j} \succ Y_r^{i,j} \mid X^{i,j},  Z_i; \phi,\rho)\, p(X^{i,j}\mid Z_i; \phi,\rho)\\
\end{align}

where $r^*$ denotes the true reward function for the annotator. Let the policy parametrized as $\pi_{\phi^*, Z}$ be the policy that is optimal for the true reward function $r^*$ for the given latent type $Z$ \cite{rafailov2023direct}. Now, the probability function in the first term can also be written in closed form in terms of these policy parameters:
\begin{align}
    P_{\phi_t}(y_w \succ Y_r \mid X, Z) = \frac{\exp\!\Big(\beta \log \tfrac{\pi_{\phi_t, z_k}(y_w \mid X)}{\pi_{\text{SFT}}(y_w \mid X)}\Big)} {\sum_{y \in \{y_w\} \cup Y_r} \exp\!\Big(\beta \log \tfrac{\pi_{\phi_t, z_k}(y \mid X)}{\pi_{\text{SFT}}(y \mid X)}\Big)}
\end{align}
where $\pi_{\phi^*,z}$ optimizes the type specific regularized objective:
\begin{align}\label{rlhf_pol_llhood:hetero}
 \pi_{\phi^*,z} = \argmax_{\pi} \E_{x \sim \mathcal{D}, y \sim \pi(y|x)}[r^*(y,x,z)] - \beta \mathbb{D}_{\text{KL}}[\pi(y|x) || \pi_{\text{SFT}}(y|x)]
\end{align}
For concise writing we introduce the following shorthand notation:
\begin{align}
    P_{\phi}(V, Z) = P_{\phi_t}(y_w \succ Y_r \mid X, Z)
\end{align}
Thus a parameterization of the policy space $\pi_{\phi,Z}$, implies a parameterization of the likelihood:
\begin{align}
    p(V_{i}\mid Z_i; \phi,\rho) = \prod_{j=1}^m P_{\phi}(V_{i,j},Z_i)\, p(X_{i,j}\mid Z_i;\phi,\rho)
\end{align}

Therefore, the M-step involves solving the following two optimization problems:
\begin{equation}
    \begin{aligned}
    \eta_{t+1} =~& \argmax_{\eta} \E_{Z\sim  p(\cdot\mid V,\theta_t)}\left[\sum_{i=1}^n \log\left(\sum_{k=1}^K \eta_k 1\{Z_i=z_k\}\right)\right]\\
    \phi_{t+1} =~& \argmax_{\phi,\rho} \E_{Z\sim p(\cdot\mid V,\theta_t)}\left[\sum_{i=1}^n \sum_{j=1}^m \log(P_{\phi}(V_{i,j},Z_i))\, + \sum_{i=1}^n \sum_{j=1}^m \,\log(p(X_{i,j}\mid Z_i;\rho))\right] \\
\end{aligned}
\end{equation} 

The first optimization problem for $\eta_{t+1}$ admits a closed-form solution. Letting $w_{k,t} = \sum_{i=1}^n p(z_k\mid V_i;\theta_t)$
\begin{align}
\E_{Z\sim p(\cdot\mid V,\theta_t)}\left[\sum_{i=1}^n \log\left(\sum_{k=1}^K \eta_k 1\{Z_i=z_k\}\right)\right] =~& \sum_{i=1}^n \sum_{k=1}^K p(z_k\mid V_i; \theta_t) \log(\eta_k) \\=~& \sum_{k=1}^K w_{k,t} \log(\eta_k)
\end{align}
Thus the optimization problem that determines $\eta_{t+1}$ takes the simple form $\max_{\eta\in \Delta(K)} \sum_{k=1}^K w_{k,t} \log\left(\eta_k\right)$.
The Lagrangian of this problem is $L(\eta, w_t, \lambda) = \sum_{k=1}^K w_{k,t} \log(\eta_k) + \lambda^T(\eta - 1)$.
The KKT condition is:
\begin{align}
    ~&\frac{w_{k,t}}{\eta_{k,t+1}} = \lambda \implies \eta_{k,t+1} \propto w_{k,t} \implies \eta_{k,t+1} = \frac{w_{k,t}}{\sum_{k} w_{k,t}}
\end{align}

Moreover, since $\sum_{k} p(z_k\mid V_i; \theta_t) = 1$, we have $\sum_{k} w_{k,t}=n$. Thus, the above simplifies to:
\begin{align}
    \eta_{k,t+1} = \frac{1}{n} w_{k,t} = \frac{1}{n} \sum_{i=1}^n p(z_k\mid V_i;\theta_t)
\end{align}
For the second optimization problem for $\phi_{t+1}$, assuming that the parameter $\rho$ that determines that $p(X\mid Z;\rho)$ is not subject to joint constraints with the parameter $\phi$, we can drop the second part in the objective to get:
\begin{equation}
    \begin{aligned}
    \phi_{t+1} 
    =~& \argmax_{\phi} \E_{Z\sim p(\cdot\mid V,\theta_t)}\left[\sum_{i=1}^n \sum_{j=1}^m \log(P_{\phi}(V_{i,j},Z_i)) \right]\\
    =~& \argmax_{\phi} \sum_{i=1}^n \E_{Z_i\sim p(\cdot\mid V_i;\theta_t)}\left[\sum_{j=1}^m \log(P_{\phi}(Z_i, V_{ij})) \right]
\end{aligned}
\end{equation} 

\subsubsection{E-Step}
The only remaining quantity we need to compute $Q(\theta\mid \theta_t)$ is the posterior distribution $p(Z\mid V;\theta)=\prod_{i=1}^n p(Z_i\mid V_i;\theta)$ for any given parameter $\theta$. First we apply the Bayes rule and later substitute the parametric form of $p(V|Z;\theta)$ and $P(Z;\theta)$ that we derived in the M-Step to get:
\begin{align}
p(z_k\mid V_i;\theta) =~& \frac{p(V_i, z_k; \theta)}{p(V_i;\theta)} \\
=~& \frac{p(V_i\mid z_k; \theta)\, p(z_k;\theta)}{\sum_{\ell=1}^K p(V_i\mid z_\ell;\theta)\, p(z_\ell;\theta)} \\
=~& \frac{p(V_i\mid z_k; \phi)\, \eta_k}{\sum_{\ell=1}^K p(V_i\mid z_\ell;\phi)\, \eta_{\ell}}\\
=~& \frac{\prod_{j=1}^m P_{\phi}(z_k, V_{ij})\, p(X_{ij}\mid z_k;\theta)\, \eta_k}{\sum_{\ell=1}^K \prod_{j=1}^m P_{\phi}(z_\ell, V_{ij})\, p(X_{ij}\mid z_\ell;\theta)\, \eta_{\ell}}.
\end{align}

Now, we can invoke assumption \ref{ass:uncorr1} to get:
\begin{align}\label{eqn:posterior2}
    p(z_k\mid V_{i};\theta) =~& \frac{\prod_{j=1}^m P_{\phi}(z_k, V_{ij}) \rho(X_{ij})\, \eta_k}{\sum_{\ell=1}^K \prod_{j=1}^m P_{\phi}(z_\ell, V_{ij}) \rho(X_{ij})\, \eta_{\ell}}
\end{align}
Note that we can write:
\begin{align}
    &\sum_{\ell=1}^K \prod_{j=1}^m P_{\phi}(z_\ell, V_{ij})\, \rho(X_{ij})\, \eta_{\ell} \\=~& \sum_{\ell=1}^K \prod_{j=1}^m \rho(X_{ij}) \cdot \prod_{j=1}^m P_{\phi}(z_\ell, V_{ij})\eta_{\ell} \\
    =~& \prod_{j=1}^m \rho(X_{ij}) \cdot \sum_{\ell=1}^K  \prod_{j=1}^m P_{\phi}(z_\ell, V_{ij})\eta_{\ell}
\end{align}
Thus, the terms $\prod_{j=1}^m \rho(X_j)$ cancel from the numerator and denominator in Equation~\eqref{eqn:posterior2}, leading to the simplified formula that is independent of $\rho$:
\begin{align}
     p(z_k\mid V_i;\theta) =~& \frac{\eta_k\, \prod_{j=1}^m P_{\phi}(z_k, V_j)}{\sum_{\ell=1}^K \eta_\ell\, \prod_{j=1}^m P_{\phi}(z_\ell, V_j)}
\end{align}

\subsubsection{Final Algorithm}
Now we put together the E and M step. Let $\gamma_{i,k} = p(z_k\mid V_i;\theta_t)$, now,
\begin{align}
    \gamma_{i,k} &= \frac{\eta_k\, \prod_{j=1}^m P_{\phi}(z_k, V_{i,j})}{\sum_{\ell=1}^K \eta_\ell\, \prod_{j=1}^m P_{\phi}(z_\ell, V_j)}\\
    \eta_{k,t+1} &= \frac{1}{n} \sum_{i=1}^n \gamma_{i,k}\\
    \phi_{t+1} &= \argmax_{\phi} \sum_{i\in {\cal I}} \sum_{k=1}^K \gamma_{i,k} \sum_{j=1}^{m_i} \log(P_{\phi}(V_{i,j}, z_k,))
\end{align}

\section{Min-Max Regret Aggregation Algorithm Variants}
The algorithm that directly minimizes the regret objective provided in ~\ref{minmax-reg-obj} is shown in Algorithm \ref{mmra_original}. This algorithm requires generation and scoring from the training policy as well as scoring from all the ensemble policies at every step of training, which consumes both compute and memory. 

Algorithm \ref{mmra_affine} is another computationally light-weight algorithm. This algorithm constrains the search space of the trained algorithm to the affine space of the ensemble policies. This doesn't require re-training a new policy and just involved selecting an ensemble among the already trained policies. As such, we define the ensemble space of policies as:
\begin{align*}
    \Pi = \left\{\sum_{k=1}^K w_k \pi_{\phi, z_k}: w \in \Delta(K)\right\}
\end{align*}
The min-max regret optimization problem can be written as a function of the policies trained using EM-DPO  without any access to the explicit reward functions, as it is a difference in rewards (\cite{rafailov2023direct}):
\begin{align*}
\min_{w\in \Delta(K)} \max_{z\in \{z_0, z_1,\ldots, z_K\}} \sum_{k=1}^K w_k\cdot \left({\cal L}_{z,z} - {\cal L}_{z, z_k}\right)
\end{align*}
where
\begin{align*}
{\cal L}_{z,z'} := 
\begin{cases}
0 & \text{if } z = z_0, \\
\E_{x\sim {\cal D},\, y\sim \pi_{z'}^*(\cdot\mid x)}\left[\log \left( \frac{\pi_z^*(y\mid x)}{\pi_{\text{SFT}}(y\mid x)}\right)\right] & \text{otherwise}.
\end{cases}
\end{align*}

Letting ${\cal R}$ denote the $(K+1)\times K$ matrix whose $(k,k')$ entry (for $0 \leq k \leq K, 1 \leq k' \leq K$) corresponds to ${\cal R}_{k,k'} := {\cal L}_{z_k, z_k} - {\cal L}_{z_k, z_{k'}},$ we can re-write the above objective as:
\begin{align}
    \min_{w\in \Delta(K)} \max_{p\in \Delta(K+1)} p^\top {\cal R} w 
\end{align}
This is simply a finite action zero-sum game, where the minimizing player has $K$ actions and the maximizing player has $K+1$ actions. A large variety of methods can be utilized to calculate an equilibrium of this zero-sum game and hence identify the minimax regret optimal mixture weights $w^*$. For instance, we can employ optimistic Hedge vs. optimistic Hedge dynamics, which are known to achieve fast convergence rates in such finite action zero-sum games (\cite{rakhlin2013optimization}) and then use the average of the solutions over the iterates of training. This is formalized in \ref{mmra_affine}  and the solution $\pi_*$ returned constitutes a $O(\log(K)\,\log(T)\,T^{-1})$-approximate solution to the min-max regret problem (a direct consequence of the results in \cite{rakhlin2013optimization}). This completes our overall direct preference optimization procedure with unobserved heterogeneous preferences.

\begin{algorithm*}[t]
\caption{MMRA (Original)}\label{mmra_original}
\begin{algorithmic}[1]
    \State \textbf{Input:} EM-DPO ensemble policies $\{\pi^*_1,\ldots,\pi^*_K\}$; SFT policy ${\pi_{\text{SFT}}}$; prompt dataset $\mathcal{D}_x$; preference dataset $\mathcal{D}$; regularization parameter $\beta$; learning rate $\eta$
    \State \textbf{Initialize:} Current policy $\pi^0 = \pi_{\text{SFT}}$ for $t=0$; weights $(w_1^0,\ldots,w_K^0) = (1/K,\ldots,1/K)$; 
    \State \textbf{Precompute:} For each $k \in [K]$:
    \Statex \hspace{\algorithmicindent} Generate dataset $\mathcal{D}_k = \{(x, y) : x \sim \mathcal{D}_x, y \sim \pi^*_k(\cdot|x)\}$
    \Statex \hspace{\algorithmicindent} Approximate $\mathbb{E}_{x, y \sim \pi^*_k}\left[\beta \log\frac{\pi^*_k(y|x)}{\pi_{\text{SFT}}(y|x)}\right] \approx \frac{1}{|\mathcal{D}_k|}\sum_{(x,y) \in \mathcal{D}_k} \beta \log\frac{\pi^*_k(y|x)}{\pi_{\text{SFT}}(y|x)}$
    \For{$t = 0,1,\ldots,T$}
        \State Generate $\mathcal{D}^t = \{(x, y) : x \sim \mathcal{D}_x, y \sim \pi^t(\cdot|x)\}$
        \State Compute $\log \pi(y|x)$ and $\log \pi^*_k(y|x)$ for $x,y \sim \mathcal{D}^t$ and $\forall k \in [K]$
        \State Approximate: 
        \State \hspace{\algorithmicindent} $\mathbb{E}_{x, y \sim \pi^t}\left[\beta \log\frac{\pi^*_k(y|x)}{\pi_{\text{SFT}}(y|x)}\right] \approx \frac{1}{|\mathcal{D}^t|}\sum_{x,y \in \mathcal{D}^t} \beta \log\frac{\pi^*_k(y|x)}{\pi_{\text{SFT}}(y|x)}$, $\forall k \in [K]$
        \State \hspace{\algorithmicindent} $\E_{x, y \sim \pi^t}\left[\beta \log\frac{\pi^t(y|x)}{\pi_{\text{SFT}}(y|x)} \right] \approx \frac{1}{|\mathcal{D}^t|}\sum_{x,y \in \mathcal{D}^t} \beta \log\frac{\pi^t(y|x)}{\pi_{\text{SFT}}(y|x)}$
        \State Compute the loss $\mathcal{L}^t_k =  w^t_k ([R_k(\pi^t)]^+  + \beta \mathbb{D}_{KL}(\pi^t || \pi_{\text{SFT}}))$.
        \State $\pi^t, w^t_k \leftarrow$ (regular or optimistic) GD and MWU on $\sum_k \mathcal{L}^t_k$
    \EndFor
\end{algorithmic}
\end{algorithm*}

\begin{algorithm}[t]
\caption{MMRA-AE (Affine Ensemble Variant)}\label{mmra_affine}
\begin{algorithmic}[1]
    \State \textbf{Input:} Distribution ${\cal D}$ of contexts $x$.
    \State \textbf{Input:} Population-specific optimal policies $\pi_{z}^*$ returned from EM-DPO
    \State \textbf{Input:} Number of iterations $T$ and a sufficiently small, albeit constant, independent of $T$, step-size $\eta$
    \State Calculate discrepancies for $z,z'\in \{z_1,\ldots, z_k\}$:
    $$
    {\cal L}_{z,z'} := \E_{x\sim {\cal D}, y\sim \pi_{z'}^*(\cdot\mid x)}\left[\log \left( \frac{\pi_z^*(y|x)}{\pi_{\text{SFT}}(y|x)}\right)\right]
    $$
    with the convention that ${\cal L}_{z_0,z_0}={\cal L}_{z_0, z_k}=0$
    \State Calculate $(K+1)\times K$ regret matrix ${\cal R}$, whose $k\in \{0,\ldots, K\}$ and $k'\in \{1,\ldots, K\}$ entry is:
    $$
    {\cal R}_{k,k'} :=  {\cal L}_{z_k, z_k} - {\cal L}_{z_k, z_k'} 
    $$
    \State Initialize $w_0 = (1/K,\ldots, 1/K)$ and $p_0=(1/(K+1),\ldots, 1/(K+1))$
    \For{$t$ in $\{0, \ldots, T\}$}
        $$
            w_t \propto w_{t-1} \exp\left\{-\eta\cdot \left(2{\cal R}^\top p_{t-1} - {\cal R}^\top p_{t-2}\right)\right\}
        $$
        $$
        p_t \propto p_{t-1} \exp\left\{\eta\cdot \left(2{\cal R} w_{t-1} - {\cal R} w_{t-2}\right)\right\}
        $$
    \EndFor
    \State \textbf{Return:} Policy $\pi^* = \sum_{k=1}^K w^*(k) \pi_{z_k}^*$, where $w^*=\frac{1}{T}\sum_{t=1}^T w_{t}$ and $w(k)$ is the value of the $k^{th}$ co-ordinate of $w$.
\end{algorithmic}
\end{algorithm}

%% file: sections/appendix_experiment.tex
\section{Additional Experiment Details}\label{appendixE}

\subsection{Data Generation}\label{appendixDataGen}

\textbf{Global Opinion QA:} This dataset contains country-level polling data on politics, religion, economics, and related topics. For each question, annotators respond according to their country-specific distribution, after which a second option is randomly rejected. For instance, a Mexican annotator answering “Do you support or oppose using the army to fight drug traffickers?” has probabilities of 84\% support, 13\% oppose, and 3\% refuse/don’t know. We focus on four countries: Britain, Indonesia, Mexico, and Pakistan, comprising a total of X questions. Each question is expanded with 10 GPT-4-generated rephrasings, yielding 11 variants per question (original + 10 rephrasings).

For model training, eight variants are used, while three are reserved for validation and testing. Annotators are simulated according to the following country proportions: 15\% from Britain, 20\% from Indonesia, 30\% from Mexico, and 35\% from Pakistan. To construct a user, 32 unique questions are sampled and a random rephrasing from the appropriate split is assigned for each. Using this approach, we generate 1,500 users for training and 400 for testing.\\

\textbf{MPI:} The MPI dataset contains 990 unique phrases, each scored $-1$, $0$, or $+1$ on one of the five OCEAN traits (Openness, Conscientiousness, Extraversion, Agreeableness, Neuroticism). For example, ``act wild and crazy'' scores $+1$ on Extraversion, while ``readily overcome setbacks'' scores $-1$ on Conscientiousness.  

We define three synthetic personalities as vectors in $\mathbb{R}^5$: $P1 = (3, 0, 2, 0, -2.5)$, $P2 = (-3, 0, -2, 0, 2.5)$, and $P3 = (0, 2, 0, 2, 0)$. sampled with probabilities $0.3$, $0.3$, and $0.4$. Note that $P2 = -P1$, creating an adversarial pair that is non-identifiable under binary preferences. Phrase rewards are computed as the inner product of the personality vector with the phrase’s trait scores. For instance, if a phrase scores $+1$ on Openness, $P1$ assigns a reward of $3 \times 1 = 3$.  

Data are generated as follows: (i) sample a personality; (ii) select one of 50 paraphrases of the instruction ``Choose the option that resonates most with your personality''; (iii) draw phrases from the MPI dataset; and (iv) simulate preferences using the Bradley--Terry model with either two or three items. Each user contributes exactly one preference pair, preventing identifiability beyond the conditions stated in Lemma~X.  

\subsection{Cluster DPO}\label{appendixClusterDPO}

The Cluster-DPO policy is generated as follows: we naively cluster the  users into 2 user sub-groups using $k$-means clustering on the average embedding of all the preferred texts of that user. Embeddings are generated using the RoBERTa-Large model \cite{liu2019roberta}. Then, we train a DPO policy on each cluster separately to get an ensemble of policies; we are essentially replacing the EM-DPO step with a $k$ means clustering step in the proposed algorithm pipeline.

\subsection{Additional results}\label{appendixEMDPO-experiment}
\subsubsection{Hyperparameter Tuning} 
We don't know the exact number of latent sub-groups apriori, so we treat this as a hyper-parameter and tune it using a validation dataset. The validation set would contain extensive preference information from a small set of diverse users. In our case, we generate dataset from each of the preference group. We then use the same accuracy and reward margins metrics from the experiment section. We plot this in the given table. While the accuracy metric doesn't observe any trends, we observe that the reward margins either peak at $k=4$ or increase only marginally for $k > 4$ across validation data from any of the component sub-groups for both datasets. Therefore, regardless of the composition of the validation dataset (which is unobservable in practice due to the latent nature of the groups) $k=4$ emerges as the optimal hyperparameter for this experiment and the two given datasets.

In practical scenarios, in addition to using a validation dataset, we must consider resource availability as well for fixing the number of groups. The memory and compute requirements of EM-DPO scale linearly and at least linearly, respectively, with the number of latent groups. The latter scaling depends on the number of steps required for convergence, which may increase with the number of groups. 

\begin{figure}[H]
\centering
\begin{subfigure}[b]{0.48\textwidth}
    \centering
    \includegraphics[width=\textwidth]{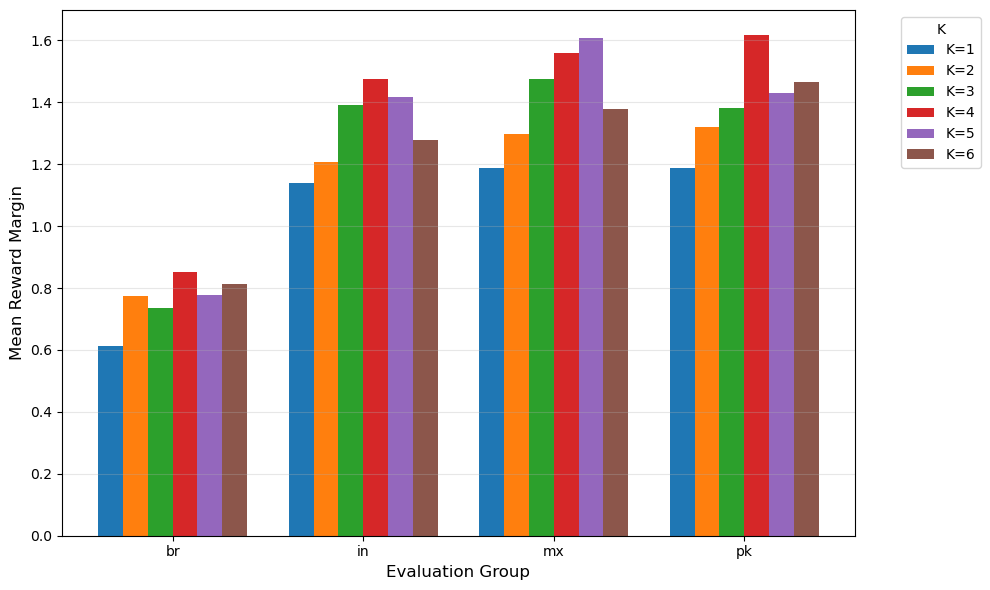}
    \caption{Global Opinion Margin}
    \label{fig:go-margin}
\end{subfigure}
\hfill
\begin{subfigure}[b]{0.48\textwidth}
    \centering
    \includegraphics[width=\textwidth]{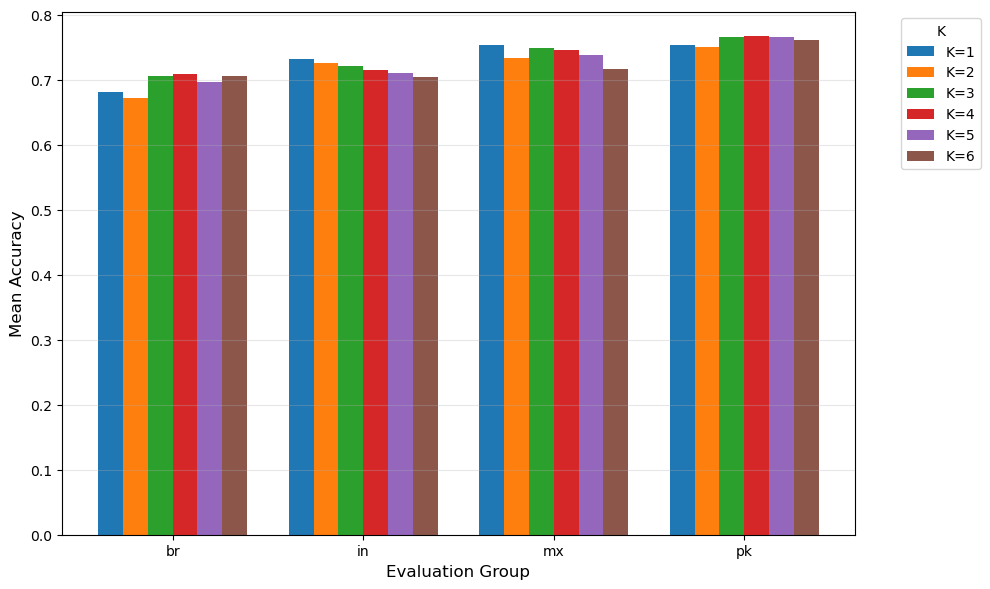}
    \caption{Global Opinion Accuracy}
    \label{fig:go-accuracy}
\end{subfigure}

\vspace{0.5cm}

\begin{subfigure}[b]{0.48\textwidth}
    \centering
    \includegraphics[width=\textwidth]{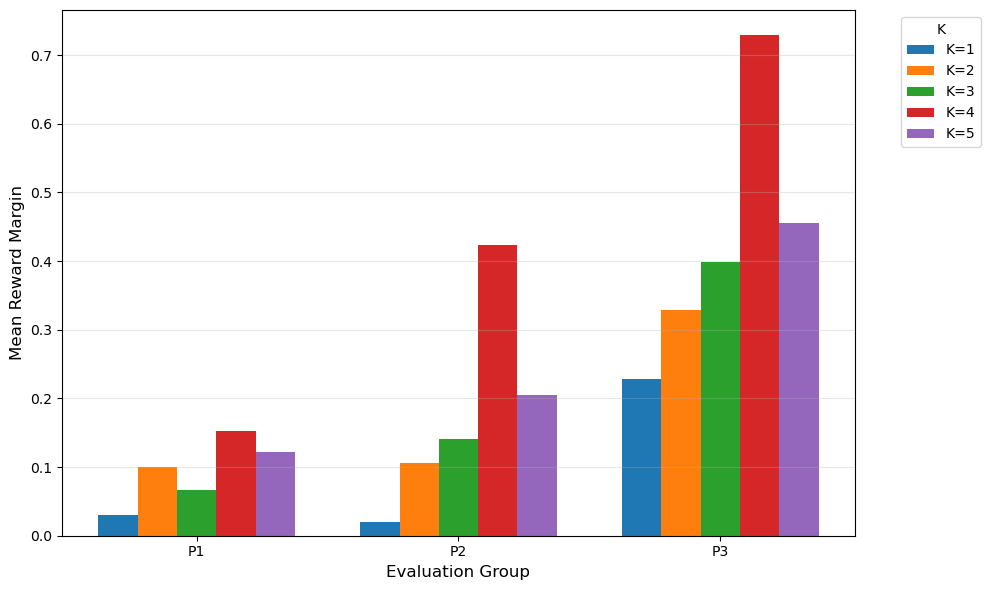}
    \caption{MPI Margin}
    \label{fig:mpi-margin}
\end{subfigure}
\hfill
\begin{subfigure}[b]{0.48\textwidth}
    \centering
    \includegraphics[width=\textwidth]{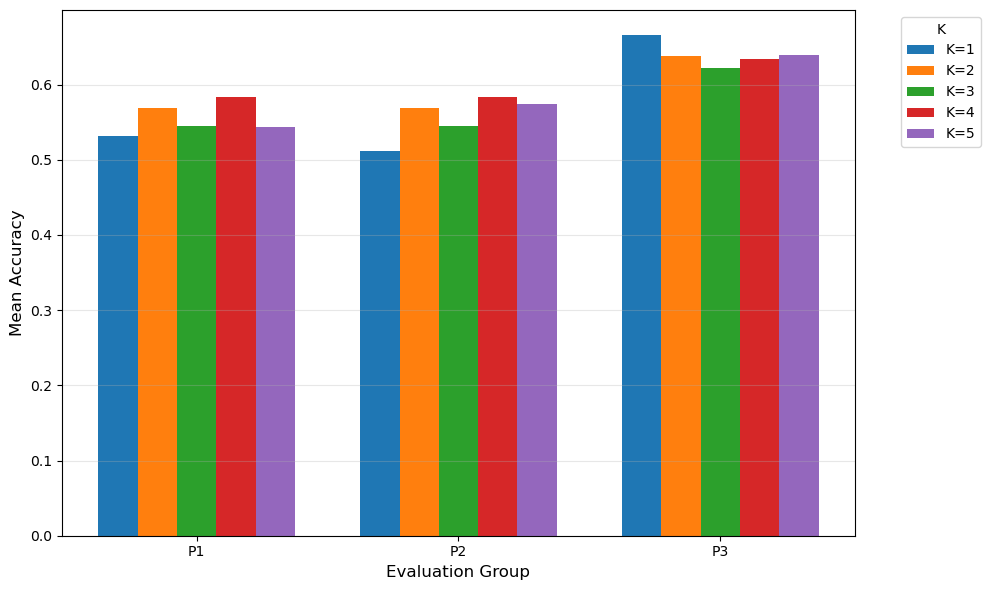}
    \caption{MPI Accuracy}
    \label{fig:mpi-accuracy}
\end{subfigure}

\caption{Hyper-parameter Tuning Results}
\label{fig:hyper-param-tuning}
\end{figure}


\subsection{Hyperparameters}

The following tables shows the hyperparameters for the LLM experiment. We ran the experiments on 8 NVIDIA H100 GPUs with 80GB memory per GPU. On average 1 run of DPO took about 15 mins for the EMDPO algorithm. For the MMRA algorithm, it took roughly 1 hour for the first generation and acoring phase, following by 15 mins for each iteration of the training \& evaluation phase.

\begin{table*}[ht]
\centering
\begin{tabular}{|l|l|}
\hline
\multicolumn{1}{|c|}{\textbf{Parameter}}           & \multicolumn{1}{c|}{\textbf{Value}} \\ \hline
Base Model                                         & Mistral 7B v0.3                     \\ \hline
Batch Size                                         & 4                                   \\ \hline
Evaluation Batch Size                              & 16                                  \\ \hline
Learning Rate                                      & 5e-7                                \\ \hline
Gradient Accumulation Steps                        & 1                                   \\ \hline
Max Gradient Norm                                  & 10.0                                \\ \hline
Max Text Length (Prompt + Response)                & 512                                 \\ \hline
Max Prompt Length                                  & 256                                 \\ \hline
No. of Training Epochs                             & 1                                   \\ \hline
No. of Evaluation Examples                         & 256                                 \\ \hline
Optimizer                                          & RMSprop                             \\ \hline
No. of Warmup Steps for Learning Rate              & 150                                 \\ \hline
No. of Iterations of the EM Algorithm              & 5                                  \\ \hline
DPO Beta                                           & 0.1                                 \\ \hline
\end{tabular}
\caption{Hyperparameters for EMDPO}
\label{tab:hyperparameters-emdpo}
\end{table*}

\begin{table*}[ht]
\centering
\begin{tabular}{|l|l|}
\hline
\multicolumn{1}{|c|}{\textbf{Parameter}}           & \multicolumn{1}{c|}{\textbf{Value}} \\ \hline
Base Model                                         & Mistral 7B v0.3                     \\ \hline
Batch Size                                         & 32                                  \\ \hline
Evaluation Batch Size                              & 32                                  \\ \hline
Learning Rate for Policy                           & 5e-7                                \\ \hline
Gradient Accumulation Steps                        & 1                                   \\ \hline
Max Gradient Norm                                  & 10.0                                \\ \hline
Max Text Length (Prompt + Response)                & 512                                 \\ \hline
Max Prompt Length                                  & 256                                 \\ \hline
No. of Prompts for Generation                      & 3,072                               \\ \hline
Completions per Prompt                             & 1                                   \\ \hline
Optimizer                                          & RMSprop                             \\ \hline
No. of MWU Iterations                              & 20                                 \\ \hline
Batches per MWU Iteration                          & 250                                 \\ \hline
MWU Learning Rate ($\eta$)                         & 0.01                                \\ \hline
\end{tabular}
\caption{Hyperparameters for MMRA (Lightweight)}
\label{tab:hyperparameters-regret-dpo}
\end{table*}